\documentclass[letterpaper, 10 pt, conference]{ieeeconf}  

\IEEEoverridecommandlockouts                              

\usepackage{graphics} 
\usepackage{epsfig} 
\usepackage{mathptmx} 
\usepackage{times} 
\usepackage{amsmath} 
\usepackage{amssymb}  
\usepackage[pagebackref,breaklinks,colorlinks,citecolor=cvprblue]{hyperref}

\usepackage{multirow}
\usepackage{booktabs}
\usepackage{cleveref}
\usepackage{stfloats}

\Crefname{figure}{Fig.}{Figs.}

\DeclareMathAlphabet{\mathcal}{OMS}{cmsy}{m}{n}

\title{\LARGE \bf
GlobalMapNet: An Online Framework for Vectorized Global HD Map Construction}

\author{Anqi Shi$^{1}$, Yuze Cai$^{1}$, Xiangyu Chen$^{1}$, Jian Pu$^{2}$, Zeyu Fu$^{3}$ and Hong Lu$^{1}$
\thanks{$^{1}$Anqi Shi, Yuze Cai, Xiangyu Chen, and Hong Lu
are with Shanghai Key Laboratory of Intelligent Information Processing, School of Computer Science, Fudan University, Shanghai 200433, China. E-mail: \{aqshi22, 24210240113, 24210240004\}@m.fudan.edu.cn; honglu@fudan.edu.cn}
\thanks{$^{2}$Jian Pu is with Institute of Science and Technology for Brain-Inspired Intelligence, Fudan University, Shanghai, China. E-mail: jianpu@fudan.edu.cn}
\thanks{$^{3}$Zeyu Fu is with Department of Computer Science, University of Exeter. E-mail: z.fu@exeter.ac.uk}
\thanks{(Corresponding author: Hong Lu.)}
}

\begin{document}

\maketitle
\thispagestyle{empty}
\pagestyle{empty}

\begin{abstract}
High-definition (HD) maps are essential for autonomous driving systems. Traditionally, an expensive and labor-intensive pipeline is implemented to construct HD maps, which is limited in scalability. In recent years, crowdsourcing and online mapping have emerged as two alternative methods, but they have limitations respectively. In this paper, we provide a novel methodology, namely global map construction, to perform direct generation of vectorized global maps, combining the benefits of crowdsourcing and online mapping. We introduce GlobalMapNet, the first online framework for vectorized global HD map construction, which updates and utilizes a global map on the ego vehicle. To generate the global map from scratch, we propose GlobalMapBuilder to match and merge local maps continuously. We design a new algorithm, Map NMS, to remove duplicate map elements and produce a clean map. We also propose GlobalMapFusion to aggregate historical map information, improving consistency of prediction. We examine GlobalMapNet on two widely recognized datasets, Argoverse2 and nuScenes, showing that our framework is capable of generating globally consistent results.
\end{abstract}    
\section{Introduction}
\label{sec:intro}

High-definition (HD) maps are highly accurate maps that provide detailed road information, such as geometric features of road boundaries, lanes, and pedestrian crossings. For high-level autonomous vehicles, HD map is crucial for accurate localization \cite{wang2021visual, petek2022robust}, which forms the basis of safe autonomous driving. However, traditional HD map production requires expensive mobile mapping systems (MMSs) and excessive human labor, making it difficult to maintain up-to-date maps in a large scale \cite{elghazaly2023high, zhuy2024ecosystem}. 

Recent works facing up to this challenge can be divided into two categories: offline HD map crowdsourcing and online HD map construction (online mapping). Crowdsourcing methods utilize sensor data generated from massive vehicles \cite{liu2024high, guo2024review}, which is adequate and cheap. Collected data are automatically preprocessed by cloud services, while manual labeling is not fully omitted \cite{dabeer2017end, kim2021updating, ye2023recruiting}. On the other hand, online mapping alleviates the burden of laborious stages \cite{guo2024review, tang2023high}, which directly predicts a local map from the surrounding environment on the ego vehicle. However, it is challenging to produce temporal consistent results. Also, former methods \cite{hdmapnet, superfusion, yuan2024streammapnet, chen2024maptracker} are not able to generate vectorized global maps like crowdsourcing does.

To move a step forward, we emphasize the \textbf{Static Map Assumption}, which means from the global perspective, the ground-truth map remains unchanged in a certain period of time, regardless of illumination, weather, and pose change of sensors. Therefore, we combine crowdsourcing with online mapping, bringing an online framework that performs closed-loop vectorized global map construction and utilization, to produce globally consistent results. It is possible to incorporate multi-run perception results from massive vehicles, and the vectorized map is space-saving for practical use on ego vehicles. The detailed comparison is shown in \Cref{tab:method_compare}.

\begin{table*}
  \centering
  \caption{Comparison between the four methods mentioned in Section 1.}
  \begin{tabular}{@{}lcccc@{}}
    \toprule
    Method & Data Collection & Location of Computation & Degree of Automation & Output \\
    \midrule
    Traditional Pipeline & MMS & Cloud Service & Rarely Automated & Global Map \\
    Crowdsourcing & Massive Vehicles & Cloud Service & Partially Automated & Global Map \\
    Online Local Map Construction & Single Vehicle & Ego Vehicle & Fully Automated & Local Map \\
    Online Global Map Construction (Ours) & Massive Vehicles & Ego Vehicle & Fully Automated & Global Map \\
    \bottomrule
  \end{tabular}
  \label{tab:method_compare}
   \vspace{-0.5cm}
\end{table*}

\begin{figure}[t]
  \centering
   \includegraphics[width=1.0\linewidth]{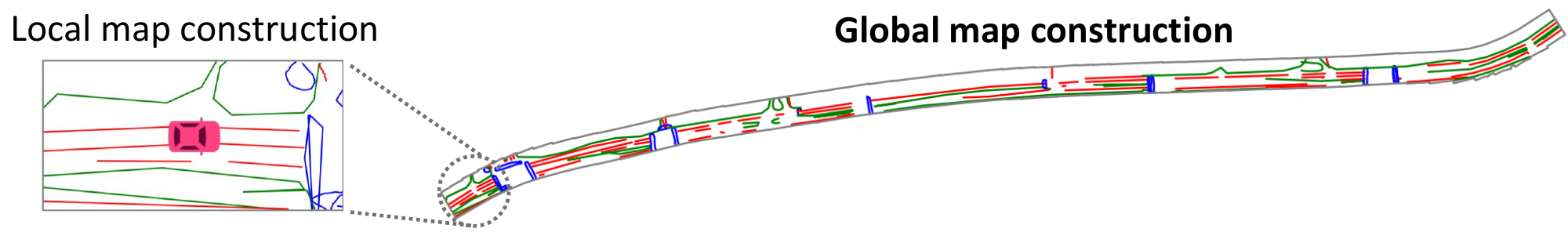}
   \caption{The relationship and difference between local map construction and global map construction. In global map construction, multi-run local mapping results are merged sequentially to produce the global map.}
   \label{fig:cross_scene}
   \vspace{-0.5cm}
\end{figure}

In this paper, we present \textbf{GlobalMapNet}, an online framework for vectorized global HD map construction. Based on local mapping methods, GlobalMapNet keeps an extraordinary global map as the long-term memory, which is obtained by continuously merging local perception results as depicted in \Cref{fig:cross_scene}. Also, this global map can be rasterized and fused with bird's-eye-view (BEV) features improve real-time prediction. Our method differs from crowdsourcing methods in that, it produces vectorized map elements with an online framework, which can be directly applied to downstream tasks like localization and planning to enable multi-task knowledge exchanging in an end-to-end driving system \cite{hu2023planning}.

To summarize, this paper makes the following contributions:

\begin{itemize}
    \item We introduce the first online framework for vectorized global HD map construction, namely GlobalMapNet, with the ability to continuously update and utilize a global map, producing consistent perception results.
    \item We formulate the process of online global map construction and address major concerns on evaluation with global average precision (GAP), a novel metric designed for global map evaluation.
    \item We conduct experiments on both nuScenes and Argoverse2 datasets, and show the effectiveness of our method by examining both local and global map construction.
\end{itemize}

\section{Relate works}
\label{sec:related}
\noindent\textbf{Crowdsourcing.} Crowdsourcing aims at lowering both the cost of expensive devices and human labor. Researches are conducted on various sectors \cite{liu2024high}, including data collection and cleaning \cite{kim2021updating, crowd:datacollect}, simultaneous localization and
mapping (SLAM) \cite{orbslam3, crowd:mapping}, feature reconstruction \cite{tang2023thma}, change detection and map update \cite{crowd:lanemark, zhang2021real, crowd:hetero, crowd:update}.

Recent works address the automated generation of road structures based on crowdsourced data. \cite{traffic-flow} focuses on producing topological maps of road intersections. It collects and predicts a semantic map, together with accumulated traffic flows on the massive vehicles. On-cloud alignment is performed to form a consistent global map, using an optimization method based on Transformer \cite{vaswani2017attention}. Road intersections are detected by forming a polygon from pedestrian crossings, then traffic flows are clustered and postprocessed to generate a topology of the intersection. MapCVV \cite{mapcvv} generates vectorized maps based on semantic road elements predicted on massive vehicles.  The on-cloud system performs single-run aggregation for inter-frame consistency and multi-run aggregation for global consistency. Element-level optimization is adopted to minimize the internal error within a local subpart of the map, promoting absolute accuracy.

Crowdsourcing methods specialize in integrating multi-run data into a unified global map. However, this part is mostly done by cloud services, preventing real-time interactions with the online driving system. Our work suggests aggregation on the ego vehicle, producing a globally consistent map with an online framework.

\noindent\textbf{Online mapping with temporal modeling.} Online mapping methods directly predict a local map on the ego vehicle. A common practice is to leverage short-term temporal information. HDMapNet \cite{hdmapnet} performs temporal fusion on rasterized maps by averaging probabilities over several frames. Tesla displays in its AI Day 2021 a temporal BEV mapping system with Spatial RNN, producing consistent rasterized results. StreamMapNet \cite{yuan2024streammapnet} generates vectorized local maps with the propagated BEV feature and map queries, which are iteratively updated within a driving scene. MapTracker \cite{chen2024maptracker} views online mapping as a tracking task, utilizing strided temporal information in different historical locations.

Some works exploit long-term temporal information. NMP \cite{Xiong_2023_CVPR} builds a global BEV features map on-cloud. The ego vehicle downloads a local clip to fuse with the local BEV feature and updates the fused result to the server afterward. GNMap \cite{GNMap} aggregates multi-run generation of vectorized local maps and produces a rasterized global map. Since rasterized results are expensive to store when the map scales up, HRMapNet \cite{hrmapnet} stores a historical map with 8-bit unsigned int values. The map is utilized with BEV feature fusion and map query initialization, then updated by rasterizing vectorized map elements and simply replacing pixels.

Former methods have not considered building a vectorized global map on the ego vehicle. Rasterized maps cost a lot of memory, and pixel accumulation is not aligned with the target of predicting vectorized map elements. In this paper, we suggest that it is possible to update and utilize a vectorized global map online, which can be directly used in downstream tasks like localization and planning. The key point is to keep and arrange map elements in vectorized form, which is space-saving compared to methods based on rasterized maps.
\section{Global Map Construction}
\label{sec:formulation}
\subsection{Task Formulation}
\label{ssec:task}
\noindent\textbf{Local map construction.} Local mapping around the ego vehicle can be formulated as a procedure for generating vectorized map elements (e.g. road boundaries), which are composed of categorical labels and 2D polylines on the BEV plane \cite{liu2023vectormapnet, liaomaptr}. We define a local map $M_i$ as a collection of labeled point sequences:
\begin{align}
    M_i &= \left\{\left( c_{ij}, P_{ij} \right) \right\}_{j=1}^{N_{M_i}}, \label{eq:map_definition} \\
    P_{ij} &= \left\{\left( x_{ijk}, y_{ijk} \right) \right\}_{k=1}^{N_{P_{ij}}}, \label{eq:point_sequence}
\end{align}
where $P_{ij}$ is a 2D point sequence presenting a map element in $M_i$, and $c_{ij}$ is the categorical label of $P_{ij}$.

Suppose the vanilla model $\mathbf{F}$ get a stream of camera images $\mathcal{I} = \left\{ I_i \right\}_{i=1}^{N_{\mathcal{T}}}$ as the input during the time period $\mathcal{T} = \left\{ T_i \right\}_{i=1}^{N_{\mathcal{T}}}$. The model continuously generates a stream of local maps $\mathcal{\hat{M}} = \left\{ \hat{M}_i \right\}_{i=1}^{N_{\mathcal{T}}}$, where $\hat{M}_i = \mathbf{F} \left( I_i \right)$, using only current frame information.




\noindent\textbf{Global map construction.} Based on the Static Map Assumption that ground-truth map elements are unchanged during a certain period of time, a local clip of the optimal global map $M^*_{global}$ with ego vehicle pose $p_i$ is exactly the optimal local map $M^*_i$:
\begin{align}
    M^*_i &= \mathrm{Clip} \left( M^*_{global}, p_i \right). \label{eq:optimal}
\end{align}

\noindent This encourages us to explore global map construction. As it is impractical to predict the global map $\hat{M}_{global}$ at once, we have to continuously update it by merging local maps. At time $T_i$, our map fusion model $\mathbf{F}_{mf}$ will additionally load the propagated hidden state $H_{i-1}$ (e.g. BEV feature), and the latest global map $\hat{M}_{global, i-1}$, where the local map prior $M'_{i-1}$ is clipped. The model then predicts the current local map $M_i$, which is used to update the global map into $\hat{M}_{global, i}$. This process can be formulated as follows:
\begin{align}
    \hat{M}'_{i-1} &= \mathrm{Clip} \left( \hat{M}_{global, i-1}, p_i \right), \label{eq:clip} \\
    \hat{M}_i &= \mathbf{F}_{mf} \left( I_i, H_{i-1}, \hat{M}'_{i-1} \right), \label{eq:predict} \\
    \hat{M}_{global, i} &= \mathrm{Merge} \left( \hat{M}_{global, i-1}, \hat{M}_i, p_i \right). \label{eq:merge}
\end{align}

\noindent $\hat{M}_{global, i}$ represents the overall perception result from $T_1$ to $T_i$. It can be further utilized to measure the overall quality of online perception, uploaded to the server to be inspected and corrected, or saved to local storage and transferred to other vehicles, serving as a long-term memory for multi-run perception. Equation \eqref{eq:clip} - \eqref{eq:merge} can also be used to formulate a practical paradigm for model-based offline global map construction.

\subsection{Evaluating global map construction}
\label{ssec:eval}
The average precision (AP) metric based on Chamfer Distance, often used to measure a single-frame local map prediction in online mapping literature \cite{yuan2024streammapnet, liaomaptr}, cannot promise the overall perception quality in a certain period of time. AP cannot reflect the inconsistency of map prediction, which can bring security risks to autonomous driving systems. Also, given by the Static Map Assumption formulated with \eqref{eq:optimal}, if the model produces a high-quality global map, local map predictions are more trustworthy. The reasons above derive the necessity of a global map evaluation metric.

\noindent\textbf{Local map evaluation.} We first formulate local map evaluation with AP. Suppose the model produces a series of local maps $\mathcal{\hat{M}} = \left\{ \hat{M}_i \right\}_{i=1}^{N_{\mathcal{T}}}$ and a global map $\hat{M}_{global}$ within time period $\mathcal{T} = \left\{ T_i \right\}_{i=1}^{N_{\mathcal{T}}}$, AP is given by:
\begin{align}
    AP &= \mathrm{AUC} \left( \bigcup_{i=1}^{N_{\mathcal{T}}} \mathrm{PR} \left( \hat{M}_i, M_i \right) \right),
\end{align}

\noindent where $M_i$ denotes the ground-truth local map. $\mathrm{PR}$ is the algorithm to match map elements and compute precision and recall within a pair of single-frame local maps and $\mathrm{AUC}$ computes the area under the precision-recall curve.

\noindent\textbf{Global map evaluation.} Based on AP, we derive the formulation of GAP, our novel metric for evaluating global map construction. We simply apply AP computation on $\hat{M}_{global}$ instead of $\mathcal{\hat{M}}$:
\begin{align}
    GAP &= \mathrm{AUC} \left( \mathrm{PR} \left( \hat{M}_{global}, M_{global} \right) \right).
\end{align}

\noindent Pursuing AP does not always bring better GAP, and vice versa. A framework focusing on global map construction may tolerate a little AP decrease, so long as it produces more consistent results indicated by GAP.

\begin{figure*}[t]
  \centering
   \includegraphics[width=0.82\linewidth]{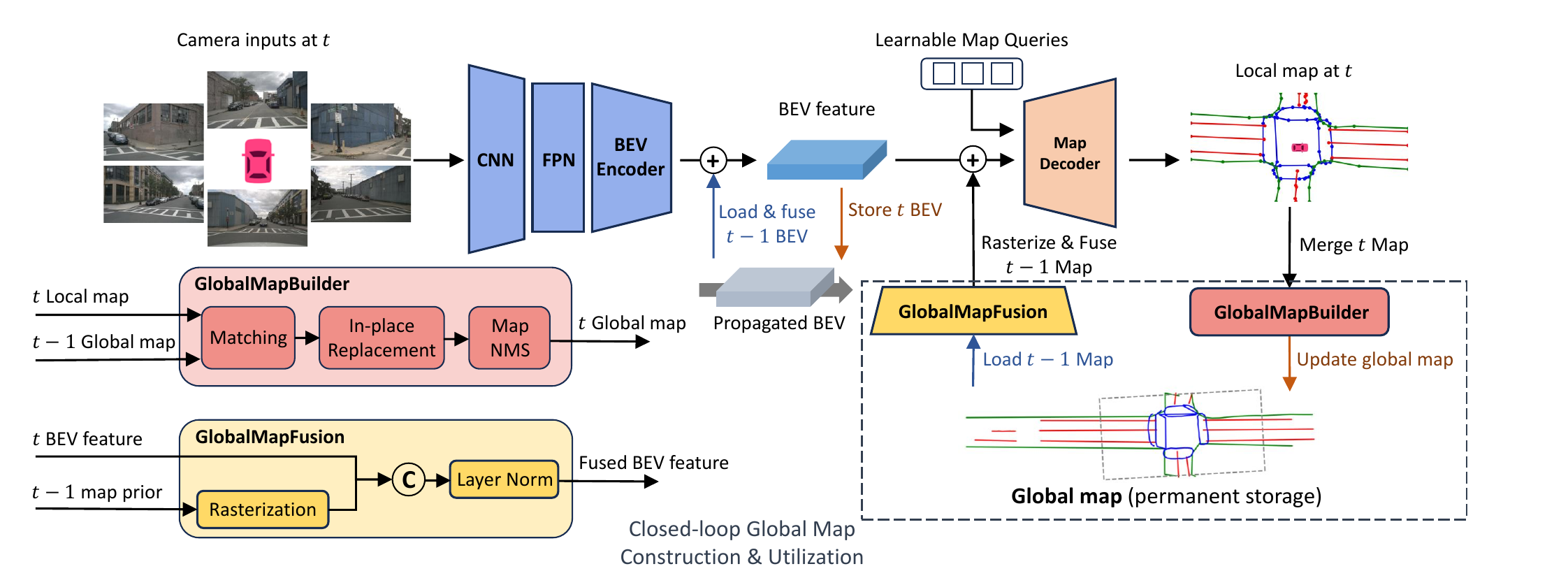}
   \caption{The structure of GlobalMapNet. Our method consists of an online local mapping system, the GlobalMapBuilder and the GlobalMapFusion. The global map is kept in permanent storage and updated continuously with local map predictions. Historical map prior is fused to produce consistent local maps, forming closed-loop global map construction and utilization. }
   \label{fig:structure}
   \vspace{-0.5cm}
\end{figure*}

\section{GlobalMapNet}
\label{sec:method}

The main idea of GlobalMapNet is to maintain and update a global map, which can be utilized as the prior for local map prediction. As shown in \Cref{fig:structure}, GlobalMapNet comprises three modules:

\begin{itemize}
    \item An \textbf{online local mapping system} that accepts sequential sensor inputs to generate local maps;
    \item The \textbf{GlobalMapBuilder} which keeps a global map memory, and continuously update it based on the latest local map prediction;
    \item The \textbf{GlobalMapFusion} module that clips a local patch from the global map, fusing historical map information and the current local feature.
\end{itemize}

\subsection{Local mapping}
\label{ssec:local_mapping}

Various online local mapping methods can fit into our global map construction framework. To keep a balance between accurate prediction and real-time computation, we choose StreamMapNet \cite{yuan2024streammapnet} as our local mapping module, which is a vision-based temporal model with simple architecture and high FPS.

\noindent\textbf{BEV feature extraction.} At first, the surrounding camera images are processed by a CNN Feature Extractor, a Feature Pyramid Network (FPN) \cite{fpn} fusion module and a BEV Encoder, producing the initial BEV feature. It is fused with historical BEV feature through a Gated Recurrent Unit (GRU) \cite{gru} network, which further propagates the fused BEV feature as a short-term memory.

\noindent\textbf{Map Decoder.} The Map Decoder is a variant of Deformable DETR \cite{zhu2020deformable}. It uses a set of learnable map queries to interact with fused BEV feature, and directly predicts map element instances in the current frame, each consists of the category and a point sequence.

\noindent\textbf{Matching and training.} The model performs a Hungarian Matching between predictions and labels, and loss is computed between matched pairs. Matching cost and loss are designed to minimize both classification error of category labeling and regression error of point sequence prediction.

Notice that we do not adopt the Query Propagation strategy in StreamMapNet, as experiments show that its effect in performance is covered by GlobalMapFusion. Also, we empirically find that historical BEV features strongly benefit local map prediction, with little extra computational and storage cost.

\subsection{GlobalMapBuilder}
\label{ssec:map_builder}
Map generation of an online local mapping system is limited to a small range. To get a global output, the GlobalMapBuilder starts with an empty global map, and continuously incorporates predicted local maps through a series of geometric algorithms, including map matching, in-place replacement and Map Non-Maximum Suppression (NMS).

\noindent\textbf{Map matching.}  At a certain frame, newly detected local map elements are transformed into global coordinates, indicating the latest perception of the global environment. A map element in this local map may be a replacement or a part of a former global map element. In that case, new and old predictions should be matched before merging.

To formulate, we define $ \left\{ P_{i}^{G} \right\}, \left\{ P_{i}^{L} \right\}$ as map elements in the global map and those in the newly predicted local map, correspondingly. Category labels are omitted, as merging only happens inside the same category. Equation \eqref{eq:clip} produces a local clip $ \left\{ P_{i}^{G^-} \right\} $ from $ \left\{ P_{i}^{G} \right\} $, which is matched with $\left\{ P_{i}^{L} \right\}$ by Hungarian Matching algorithm based on Chamfer Distance, forming matched pairs $ \left\{ \left( P_{i}^{G}, P_{j}^{L} \right) \right\}$.

\noindent\textbf{In-place replacement.} An in-place replacement strategy is adopted to merge matched pairs. A least-distance projection of $P_{i}^{L}$ onto the corresponding $P_{j}^{G}$ is computed, where sub-sequence of $P_{j}^{G}$ will be replaced by the entire $P_{i}^{L}$. Finally, we get $ \left\{ P_{i}^{G^+} \right\} $ as the merged global map, also including non-matched local and global map elements.

\noindent\textbf{Map NMS.} To further improve the quality of global map construction, we propose a novel post-processing method, namely Map NMS, to remove duplicate predictions of map elements. Similar to NMS in object detection, $ \left\{ P_{i}^{G^+} \right\} $ is first sorted by confidence score, and a map element with the higher score eliminates another if their Intersection over Union (IoU) is above the given threshold. Buffered IoU \cite{buffer} is employed to formulate the overlap between point sequences. With Map NMS, overlap within the same category can be eliminated to produce a clean global map.

\subsection{GlobalMapFusion}
\label{ssec:map_fusion}
To improve both the quality and consistency of local map prediction, the latest global map can be exploited as the prior. We employ the GlobalMapFusion module to put this idea into practice. The global map elements are first rasterized into BEV masks, then fused with the current BEV feature, which allows map queries in Map Decoder to interact with global map information.

\noindent\textbf{Soft rasterization.} For a certain category $c_i \in C$, we gather corresponding map elements into $ \left\{ P_{ij}^{G} \right\}_{j=1}^{N_{c_i}} $, where $N_{c_i}$ denotes the total amount of map elements within this category. A local clip $ \left\{ P_{ij}^{G^-} \right\}_{j=1}^{N_{c_i^-}} $ is extracted from $ \left\{ P_{ij}^{G} \right\}_{j=1}^{N_{c_i}} $. GlobalMapFusion then rasterizes these point sequences into a soft BEV mask with Gaussian-based rendering method \cite{soft-rasterizer, mapvr}:
\begin{align}
    I_{c_i} \left( x, y \right) &= \max\nolimits_{j=1}^{N_{c_i^-}} \exp \left( \frac{-D \left( x, y; P_{ij}^{G^-} \right) }{\tau} \right), \label{eq:rasterize}
\end{align}

\noindent where $I_{c_i} \left( x, y \right)$ represents the intensity of mask with category $c_i$ in position $ \left( x, y \right) $, so that $I_{c_i} \left( x, y \right) \in \left[0, 1\right)$, and $ D \left( x, y; P_{ij}^{G^-} \right) $ is the Euclidean distance between $ \left( x, y \right) $ and the point sequence $P_{ij}^{G^-}$. $\tau$ is a smoothness factor that regulates the distance, so that larger $\tau$ gives a smoother rendering.

\noindent\textbf{Utilizing traced region.} It is important for an ego vehicle to ascertain the range of the traced region (i.e. visible region) where historical perception results have covered. The traced region boundary is viewed as a special category $c_{0}$. Take this into consideration, we acquire $\lvert C \rvert + 1$ soft BEV masks $ \left\{ I_{c_i} \right\}_{i=0}^{\lvert C \rvert}$ all together.

\noindent\textbf{Fusing the historical map.} We adopt a simple yet effective way to utilize the map prior. GlobalMapFusion performs a channel concatenation between the linearly transformed BEV feature and rasterized soft BEV masks, and Layer Normalization is adopted to align these features. The fused BEV feature contains both local perception results and long-term information from the global map, both can be accessed by map queries in Map Decoder.

When there is no existing global map, all BEV masks are filled with 0. The model is trained to fully rely on the local perception inputs when there is no available map information.

\section{Experiments}

\label{sec:experiment}
\begin{table*}
  \caption{Single-scene evaluation results on nuScenes.}
  \centering
  \begin{tabular}{@{}l|l|cccc|cccc@{}}
    \toprule
    Local Mapping Range & Method 
    & $\mathrm{AP}_{road}$ & $\mathrm{AP}_{lane}$ & $\mathrm{AP}_{ped}$ & $\mathrm{mAP}$
    & $\mathrm{GAP}_{road}$ & $\mathrm{GAP}_{lane}$ & $\mathrm{GAP}_{ped}$ & $\mathrm{mGAP}$  \\
    \midrule
    \multirow{2}{*}{$60 \times 30$ m} & StreamMapNet \cite{yuan2024streammapnet}
    & 42.4 & 28.7 & 27.4 & 32.9
    & 12.8 & 13.4 & 15.5 & 13.9 \\
    & GlobalMapNet (Ours)
    & \textbf{43.4} & \textbf{31.8} & \textbf{29.3} & \textbf{34.8}
    & \textbf{18.0} & \textbf{16.3} & \textbf{18.5} & \textbf{17.6} \\
    \midrule
    \multirow{2}{*}{$100 \times 50$ m} & StreamMapNet \cite{yuan2024streammapnet}
    & \textbf{26.3} & \textbf{21.4} & \textbf{25.8} & \textbf{24.5}
    & 6.0 & \textbf{10.2} & 13.4 & 9.9 \\
    & GlobalMapNet (Ours)
    & 25.8 & 21.2 & 25.5 & 24.2
    & \textbf{6.4} & \textbf{10.2} & \textbf{20.4} & \textbf{12.3} \\
    \bottomrule
  \end{tabular}
  \\
  \vspace{0.1cm}
  \footnotesize{*``road", ``lane", ``ped" are the abbreviations for road boundary, lane divider, and pedestrian crossing, respectively.}
  \label{tab:nuscenes}
  \vspace{-0.2cm}
\end{table*}

\begin{table*}
  \caption{Single-scene evaluation results on Argoverse2.}
  \centering
  \begin{tabular}{@{}l|l|cccc|cccc@{}}
    \toprule
    Local Mapping Range & Method 
    & $\mathrm{AP}_{road}$ & $\mathrm{AP}_{lane}$ & $\mathrm{AP}_{ped}$ & $\mathrm{mAP}$
    & $\mathrm{GAP}_{road}$ & $\mathrm{GAP}_{lane}$ & $\mathrm{GAP}_{ped}$ & $\mathrm{mGAP}$  \\
    \midrule
    \multirow{2}{*}{$60 \times 30$ m} & StreamMapNet \cite{yuan2024streammapnet}
    & 64.4 & 58.5 & \textbf{58.2} & \textbf{60.4}
    & 33.8 & 34.2 & 27.2 & 31.7 \\
    & GlobalMapNet (Ours)
    & \textbf{64.8} & \textbf{58.6} & 57.5 & 60.3
    & \textbf{38.8} & \textbf{34.5} & \textbf{33.7} & \textbf{35.6} \\
    \midrule
    \multirow{2}{*}{$100 \times 50$ m} & StreamMapNet \cite{yuan2024streammapnet}
    & \textbf{52.7} & \textbf{49.2} & \textbf{61.1} & \textbf{54.3}
    & 23.0 & 25.4 & 41.3 & 29.9 \\
    & GlobalMapNet (Ours)
    & 52.1 & 47.5 & 61.0 & 53.5
    & \textbf{25.0} & \textbf{26.3} & \textbf{44.6} & \textbf{32.0} \\
    \bottomrule
  \end{tabular}
  \label{tab:argoverse2}
   \vspace{-0.5cm}
\end{table*}

We evaluate our method for both local and global map generation on two widely recognized datasets: nuScenes \cite{nuscenes2019} and Argoverse2 \cite{Argoverse2}. 

\subsection{Implementation details}
\label{ssec:implement}
\noindent\textbf{Tasks.} We base our experiments on driving scenes, each lasting for 20s (in nuScenes) or 15s (in Argoverse2), sampled at 2Hz. The inputs within each scene are a stream of surrounding camera images, 6 for nuScenes and 7 for Argoverse2, together with camera intrinsic and extrinsic parameters. The labels are the vectorized global map of this scene and a stream of vectorized local maps clipped from it. The map includes three generally concerned categories of map elements: road boundary, lane divider, and pedestrian crossing.

\noindent\textbf{Training.} We keep hyper-parameters and other training details aligned with StreamMapNet, which serves as the baseline. For both GlobalMapNet and StreamMapNet, models are trained on a single GPU with a batch size of 4 and a gradient accumulation step of 8. Each model is trained for 24 epochs, while in the first 4 epochs, GlobalMapNet keeps an empty map without update. Also, we adopt an uneven update strategy, where only 1/4 of scenes can update and fuse the stored global map frame by frame. This makes training smoother for the GlobalMapFusion module, and predicting scenes with empty maps increases the robustness of the model. 

\noindent\textbf{Evaluation.} Comparison is made to examine the effectiveness of GlobalMapFusion. Since the original StreamMapNet does not generate a vectorized global map, an identical GlobalMapBuilder is installed on it. We mainly consider mean AP (mAP) and mean GAP (mGAP) over all three categories, which indicates the overall ability of local and global map construction. The capability of GlobalMapBuilder is further explored in the ablation study and visualization.

\subsection{Results}
\label{ssec:results}
\noindent\textbf{Single-scene evaluation.} We first consider map generation within a single scene. Experiments are conducted on new train and validation splits on nuScenes and Argoverse2, to minimize location overlap \cite{yuan2024streammapnet}. Every model starts with an empty global map, updates every 4 frames (about 2 seconds) with the latest perception results, and evaluates GAP after the entire scene is traversed.

\Cref{tab:nuscenes} and \Cref{tab:argoverse2} show the comparison on nuScenes and that on Argoverse2, respectively. GlobalMapNet has an obvious superiority compared to StreamMapNet in mGAP (+3.7 for nuScenes at $60 \times 30$ m, and +3.9 for Argoverse2 at $60 \times 30$ m), suggesting that historical map prior is the key to improving global map construction. Besides, we make another two observations from the result:

\begin{enumerate}
    \item AP and GAP optimization may take different paths, as discussed in \Cref{ssec:eval}. The GlobalMapFusion module steadily benefits global map construction, but this does not guarantee a similar improvement in local map construction. 
    \item The effectiveness of GlobalMapFusion saturates to some extent under the longer-range setting. We infer that longer-range prediction forces the model to capture broader surrounding information, thus incorporating historical map information brings less improvement.
\end{enumerate}



\begin{table}
  \caption{Cross-scene evaluation results on nuScenes.}
  \centering
  \begin{tabular}{@{}l|cccc@{}}
    \toprule
    Method & $\mathrm{GAP}_{road}$ & $\mathrm{GAP}_{lane}$ & $\mathrm{GAP}_{ped}$ & mGAP \\
    \midrule
    StreamMapNet \cite{yuan2024streammapnet}
    & 6.2 & 9.9 & 15.5 & 10.5 \\
    GlobalMapNet (Ours)
    & \textbf{10.7} & \textbf{12.2} & \textbf{22.5} & \textbf{15.2} \\
    \bottomrule
  \end{tabular}
  \label{tab:cross_scene}
   \vspace{-0.3cm}
\end{table}


\noindent\textbf{Cross-scene evaluation.}
Cross-scene evaluation is a more challenging task, which examines the ability of long-term global map construction. Experiments are conducted on nuScenes, which contains scenes ranging from July 2018 to November 2018. Scenes are first sorted by timestamps as if the ego vehicle is naturally driving in order of date and time. At the first frame of every new scene, it inherits the latest historical map that contains the current position of the ego vehicle. Therefore, the range of the global map tends to grow as driving time increases, making it harder to predict long and continuous road boundaries and lane dividers.

The results are shown in \Cref{tab:cross_scene}. GlobalMapNet is still much better than StreamMapNet in mGAP (+4.7), confirming its superiority in real scenarios. The enhanced advantage in $\mathrm{GAP}_{ped}$ (+7.0) proves that GlobalMapFusion can benefit from cross-scene information in generating small map elements like pedestrian crossings consistently.

\subsection{Ablation studies}
\label{ssec:ablation}
Ablation studies are conducted on nuScenes at $60 \times 30$ m range, to analyze the effectiveness of each module, and how the parameters of GlobalMapBuilder affect global map construction.

\begin{figure*}[t]
  \centering
   \includegraphics[width=0.9\linewidth]{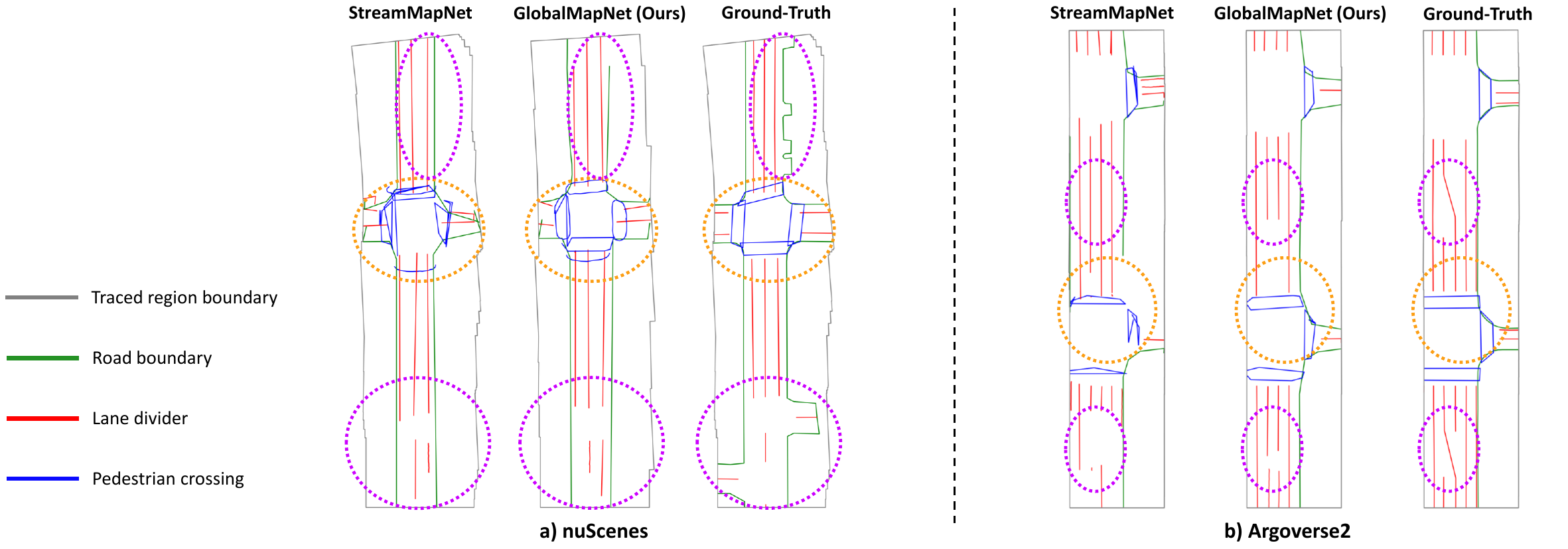}
   \caption{Visualized single-scene results on two datasets. a) nuScenes: GlobalMapNet performs better in predicting the road intersection (yellow circles). Both methods fail to generate tangled road boundaries (purple circles). b) Argoverse2: GlobalMapNet generates a continuous road boundary, while StreamMapNet fails with a broken prediction (yellow circles). Both methods fail to predict complicated lane dividers (purple circles).}
   \label{fig:results}
   \vspace{-0.5cm}
\end{figure*}
\begin{table}
  \caption{Ablation on each module.}
  \centering
  \begin{tabular}{@{}l|l|c|c@{}}
    \toprule
    Method & Module Change & mAP & mGAP \\
    \midrule
    1) StreamMapNet$-$ & Non-temporal Model
    & 30.1 & 12.6 \\
    - & $+$ BEV Feature Propagation
    & - & - \\
    2) StreamMapNet & $+$ Query Propagation
    & 32.9 & 13.9 \\
    \midrule
    - & $-$ Query Propagation
    & - & - \\
    3) GlobalMapNet$-$ & $+$ GlobalMapFusion
    & 34.0 & 17.3 \\
    \midrule
    4) GlobalMapNet
    & $+$ Traced Region Mask & \textbf{34.8} & \textbf{17.6} \\
    \bottomrule
  \end{tabular}
  \label{tab:modules}
   \vspace{-0.5cm}
\end{table}

\noindent\textbf{Ablation on each module.} Our ablation study on each module of GlobalMapNet is shown in \Cref{tab:modules}. Starting from a non-temporal model, modules are iteratively added and evaluated. Their contributions are demonstrated by mAP and mGAP increases:

\begin{enumerate}
    \item \textbf{StreamMapNet$-$} is the basic non-temporal model, which is a modified version of StreamMapNet deprived of any temporal information input.
    \item \textbf{StreamMapNet} is the original baseline model. It utilizes BEV feature propagation and Query propagation, which bring a 2.8 mAP increase and a 1.3 mGAP increase.
    \item \textbf{GlobalMapNet$-$} replaces Query propagation with GlobalMapFusion. This step brings a 1.1 mAP increase and a 3.4 mGAP increase.
    \item \textbf{GlobalMapNet} further utilizes traced region information in map fusion, which bring a 0.8 mAP increase and a 0.3 mGAP increase.
\end{enumerate}

The results indicate that GlobalMapFusion is more powerful in incorporating vectorized map prior, which is important as well as propagated BEV feature. Traced region information also benefits both local map and global map prediction, in that it can be used to tell an empty region from an unexplored region.




\begin{table}
  \caption{Ablation on parameters of GlobalMapBuilder.}
  \centering
  \tabcolsep=0.1cm
  \begin{tabular}{@{}l|ccc|cccc@{}}
    \toprule
    Method & $\mathrm{D}_{road}$ & $\mathrm{D}_{lane}$ & $\mathrm{D}_{ped}$ & $\mathrm{GAP}_{road}$ & $\mathrm{GAP}_{lane}$ & $\mathrm{GAP}_{ped}$ & $\mathrm{mGAP}$ \\
    \midrule
    \multirow{4}{*}{StreamMapNet}
    & 2.0 & 1.0 & 0.5
    & \textbf{12.8} & \textbf{13.4} & 15.5 & 13.9 \\
    & 1.0 & 1.0 & 1.0
    & 12.2 & \textbf{13.4} & \textbf{16.9} & \textbf{14.2} \\
    & 1.0 & 0.5 & 0.25
    & 12.2 & 12.1 & \textbf{16.9} & 13.8 \\
    & 4.0 & 2.0 & 1.0
    & 10.8 & 12.4 & \textbf{16.9} & 13.4 \\
    \midrule
    \multirow{4}{*}{GlobalMapNet}
    & 2.0 & 1.0 & 0.5
    & \textbf{18.0} & \textbf{16.3} & \textbf{18.5} & \textbf{17.6} \\
    & 1.0 & 1.0 & 1.0
    & 14.3 & \textbf{16.3} & 17.2 & 15.9 \\
    & 1.0 & 0.5 & 0.25
    & 14.3 & 15.2 & 14.3 & 16.6 \\
    & 4.0 & 2.0 & 1.0
    & 17.2 & 15.3 & 17.0 & 16.5 \\
    \bottomrule
  \end{tabular}
  \label{tab:params}
   \vspace{-0.5cm}
\end{table}

\noindent\textbf{Parameters of GlobalMapBuilder.} The GlobalMapBuilder should be carefully optimized to generate decent global maps. We mainly analyze the impact of two map update parameters: the chamfer distance in map matching, and the buffer distance to compute buffered IoU in Map NMS. These parameters are adjusted only at the inference stage, to merely examine the GlobalMapBuilder.

The results are shown in \Cref{tab:params}. $\mathrm{D}_{road}$, $\mathrm{D}_{lane}$ and $\mathrm{D}_{ped}$ denote the chamfer distances for road boundary, lane divider and pedestrian crossing, respectively. The buffer distances are equal to chamfer distances correspondingly. We use $\mathrm{D}_{road}=2.0$, $\mathrm{D}_{lane}=1.0$ and $\mathrm{D}_{ped}=0.5$ as the default setting, and analyze from two aspects: 1) using the same distance for different categories, and 2) scaling these parameters collectively.

We discover that these parameters can strongly affect the GAP at the inference stage. For GlobalMapNet, GAP is more sensitive to these parameters, and it's better to adjust the distance for every category according to its common pattern. For example, road boundaries are typically long and distant to each other, thus larger $\mathrm{D}_{road}$ should be adopted.

\subsection{Visualization}
\label{ssec:visualization}

To analyze single-scene performance, we examine GlobalMapNet and StreamMapNet on both nuScenes and Argoverse2 at $60 \times 30$ m range. As depicted in \Cref{fig:results}, GlobalMapBuilder helps both models to generate decent global maps with matching, replacement, and Map NMS algorithm. GlobalMapNet is superior to StreamMapNet in global map construction, showing the effectiveness of the GlobalMapFusion module. Also, predicting complex road structures remains the major challenge, as it is harder to understand long and continuous map elements with the range of the global map growing.
\section{Conclusion}
In this study, we propose GlobalMapNet to provide a novel perspective in HD map construction. Our method can practically generate vectorized global maps on massive vehicles, with efficient map building algorithms and map fusing techniques. Current global mapping framework still struggles in producing complicated road structures, especially when taking accuracy, consistency and real-time performance into account. We hope our work will facilitate future research in overcoming these difficulties.
\label{sec:conclusion}








\bibliographystyle{IEEEtran} 
\bibliography{IEEEabrv,IEEEexample}

\end{document}